\documentclass[runningheads]{llncs}
\usepackage[T1]{fontenc}
\usepackage{graphicx}
\usepackage{times}
\usepackage{soul}
\usepackage{url}
\usepackage[hidelinks]{hyperref}
\usepackage[utf8]{inputenc}
\usepackage[small]{caption}
\usepackage{graphicx}
\usepackage{amsmath}
\usepackage{booktabs}
\usepackage{algorithm}
\usepackage{algorithmic}
\usepackage{amssymb}
\usepackage{mathtools}
\usepackage{multirow}
\usepackage[american]{babel} 
\usepackage{makecell}
\usepackage {wrapfig}
\usepackage[misc]{ifsym}

\def \y {\bm{y}}
\def \x {\bm{x}}
\def \z {\bm{z}}

\def \p {\bm{p}}
\def \c {\bm{c}}
\def \a {\bm{a}}
\def \W {\bm{W}}

\def \R {\mathbb{R}}

\usepackage{bm}
\usepackage{bbding}
\usepackage{subfigure}

\begin{document}
\title{Bi-CryptoNets: Leveraging Different-Level Privacy \\ For Encrypted Inference}
\titlerunning{Bi-CryptoNets: Leveraging Different-Level Privacy for Encrypted Inference}
\author{Man-Jie Yuan \and
Zheng Zou \and
Wei Gao\textsuperscript{(\Letter)}
}
\authorrunning{M.-J. Yuan et al.}
\institute{National Key Laboratory for Novel Software Technology, Nanjing University, China \\
School of Artificial Intelligence, Nanjing University, China \\
\email{\{yuanmj,zouz,gaow\}@lamda.nju.edu.cn}
}
\maketitle              % typeset the header of the contribution
\begin{abstract}
Privacy-preserving neural networks have attracted increasing attention in recent years, and various algorithms have been developed to keep the balance between accuracy, computational complexity  and information security from the cryptographic view. This work takes a different view from the input data and structure of neural networks. We decompose the input data (e.g., some images) into sensitive and insensitive segments according to importance and privacy. The sensitive segment includes some important and private information such as human faces and we take strong homomorphic encryption to keep security, whereas the insensitive one contains some background and we add perturbations. We propose the bi-CryptoNets, i.e.,  plaintext and ciphertext branches, to deal with two segments, respectively, and ciphertext branch could utilize the information from plaintext branch by unidirectional connections. We adopt knowledge distillation for our bi-CryptoNets by transferring representations from a well-trained teacher neural network. Empirical studies show the effectiveness and decrease of inference latency for our bi-CryptoNets.

\end{abstract}
\section{Introduction}
Recent years have witnessed increasing attention on privacy-preserving neural networks \cite{Gilad-BachrachD16,BrutzkusGE19,LeeLLK0NC22}, which can be viewed as a promising security solution to the emerging Machine Learning as a Service (MLaaS) \cite{RibeiroGC15,GhodsiJRG21,GongCYMW21}. Specifically, some clients can upload their encrypted data to the powerful cloud infrastructures, and then obtain machine learning inference services; the cloud server performs inference without seeing clients' sensitive raw data by cryptographic primitives, and preserve data privacy.

Privacy-preserving neural networks are accompanied with heavy computational costs because of homomorphic encryption \cite{Gentry09,BoulemtafesDC20}, and various algorithms and techniques have been developed to keep the balance between accuracy, information security and computation complexity. For example, Brutzkus et al. proposed the LoLa network for fast inference over a single image based on well-designed packing methods \cite{BrutzkusGE19}, and Lou and Jiang introduced the circuit-based network SHE with better accuracies \cite{Lou019}, implemented with the TFHE scheme \cite{chillotti2020tfhe}. Dathathri et al. presented the compiler CHET to optimize data-flow graph of HE computations for privacy-preserving neural networks \cite{DathathriS0LLMM19} .
Yin et al. presented a comprehensive survey on privacy-preserving networks \cite{YinZH21}.

\begin{wrapfigure}{r}{6cm}
\centering
\includegraphics[width=5.8cm]{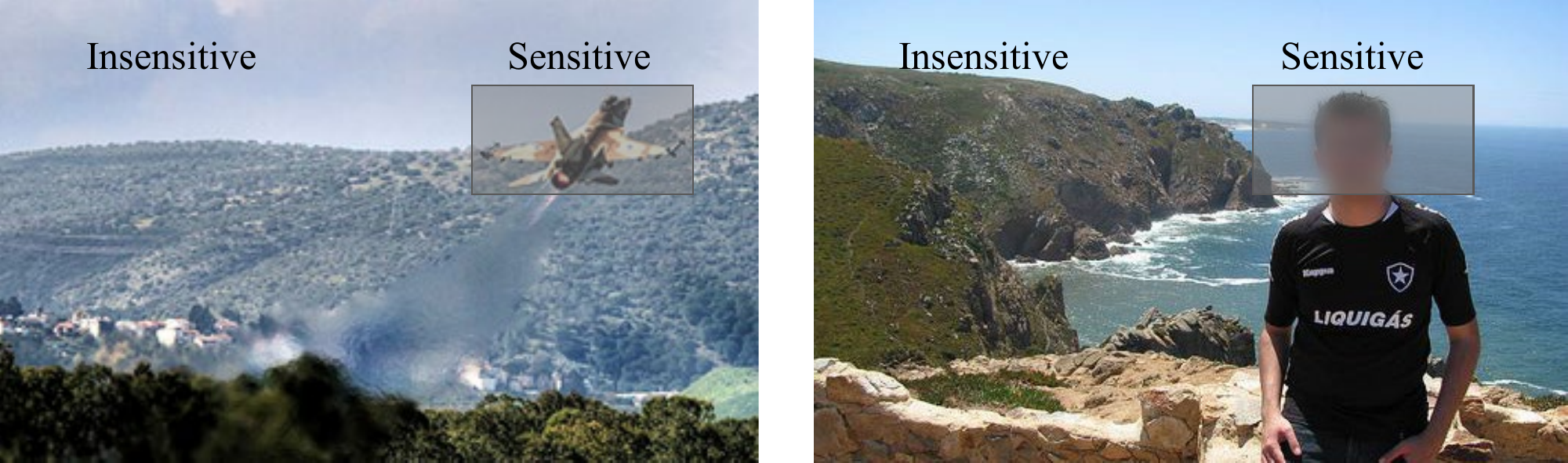}
\caption{An illustration for input data (e.g., some images\protect\footnotemark), which consists of two segments  with different importance and privacy.}
\label{fig:demo}
\vskip -0.4in
\end{wrapfigure}

Previous studies mostly encrypted the entire input data and treated indiscriminately. In some applications, however, input data may consist of sensitive and insensitive segments according to different importance and privacy. As shown in Figure~\ref{fig:demo}, a fighter is more sensitive than background such as sky and mountains in a military picture, and a human face is more private than landscape in a photo.

This work presents new privacy-preserving neural network from the view of input data and network structure, and the main contributions can be summarized as follows:

\footnotetext{Download from ILSVRC dataset \cite{yang2022study}.}

\begin{itemize}
\item We decompose input data (e.g., images) into sensitive and insensitive segments according to importance and privacy. We adopt strong homomorphic encryption to keep the security of  sensitive segment, yet mingle some perturbations \cite{dwork2016calibrating,yang2022study} to insensitive segment. This could reduce computational overhead and perform private inference of low latency, without unnecessary encryption on insensitive segment.

\item We propose the bi-CryptoNets, i.e., ciphertext and plaintext branches, to deal with sensitive and insensitive segments, respectively. The ciphertext branch could use information from plaintext branch by unidirectional connections, but the converse direction does not hold because of the spread of ciphertexts. We integrate features for the final predictions, from the outputs of both branches.

\item We present the feature-based knowledge distillation to improve the performance of our bi-CryptoNets, from a teacher of convolutional neural network trained on the entire data without decomposition.

\item We present empirical studies to validate the effectiveness and decrease of inference latency for our bi-CryptoNets. We could improve inference accuracy by $0.2\%\sim2.1\%$, and reduce inference latency by $1.15\times\sim 3.43\times$ on a single image, and decrease the amortized latency by $4.59\times\sim 13.7\times$ on a batch of images.
\end{itemize}

The rest of this work is organized as follows: Section~\ref{background} introduces relevant work. Section~\ref{method} presents our bi-CryptoNets. Section~\ref{fkd} proposes the feature-based knowledge distillation. Section~\ref{exp} conducts experiments, and Section~\ref{conclusion} concludes with future work.

\section{Relevant Work}\label{background}

\textbf{Homomorphic encryption}. Homomorphic Encryption (HE) is a cryptosystem that allows operations on encrypted data without requiring access to a secret key \cite{Gentry09}. In addition to encryption function $E$ and decryption function $D$, HE scheme provides two operators $\oplus$ and $\otimes$ such that, for every pair of plaintexts $x_1$ and $x_2$,
\begin{eqnarray*}
D(E(x_1)\oplus E(x_2))=x_1+x_2\ ,\quad
D(E(x_1)\otimes E(x_2))=x_1\times x_2\ ,
\end{eqnarray*}
where $+$ and $\times$ are the standard addition and multiplication, respectively. Hence, we could directly perform private addition, multiplication and polynomial functions over encrypted data, by using $\oplus$ and $\otimes$ operators without knowing true values in plaintexts.

The commonly-used HE cryptosystems include CKKS \cite{CheonKKS17}, BGV \cite{brakerski2014leveled} and TFHE \cite{chillotti2020tfhe} for privacy-preserving machine learning. The BGV and TFHE schemes support integer and bits computations, while the CKKS scheme supports fixed-point computations. For CKKS, a ciphertext is an element in the ring $\mathcal{R}_p^2$, where $\mathcal{R}_p=\mathbb{Z}_p[x]/(x^N+1)$ is the residue ring of polynomial ring, with polynomial degree $N$. In addition to HE, a recent study introduced a new data encryption scheme by incorporating crucial ingredients of learning algorithm, specifically the Gini impurity in random forests \cite{xie2023gini}.
\vspace{+2mm}

\noindent\textbf{Privacy-preserving neural networks}. Much attention has been paid on the privacy-preserving neural networks in recent years. For example, Gilad-Bachrach et al. proposed the first CryptoNets to show the feasibility of private CNN inference by HE scheme \cite{Gilad-BachrachD16}. Brutzkus et al. proposed the LoLa to optimize the implementation of matrix-vector multiplication by different packing methods, and achieve fast inference for single image \cite{BrutzkusGE19}. In those studies, the ReLU activation has been replaced by squared activation because the used HE schemes merely support polynomial functions.
Lou and Jiang presented a circuit-based quantized neural network SHE implemented by TFHE scheme, which could  perform ReLU and max pooling by comparison circuits to obtain good accuracies \cite{Lou019}. Recent studies utilized the CKKS scheme to avoid quantization of network parameters for better accuracies \cite{DathathriKSDLM20,LeeLLK0NC22}.
\vspace{+2mm}

\noindent\textbf{Knowledge Distillation}. Knowledge distillation focuses on  transferring knowledge from a pretrained teacher model to a student model \cite{HintonVD15,LiLYZSLLY23}. Traditional methods aimed to match the output distributions of two models by minimizing the Kullback-Leibler divergence loss \cite{HintonVD15}. Subsequently, a variety of innovative distillation techniques have been developed. Romero et al. proposed Fitnets \cite{fitnet}, which matches the feature activations, while Zagoruyko and Komodakis suggested matching attention maps between two models \cite{ZagoruykoK17}. Gou et al. gave an extensive review on knowledge distillation \cite{gou2021knowledge}.
\vspace{+2mm}

\noindent\textbf{Threat model}. Our threat model follows previous studies on private inference \cite{BoemerCCW19,LouLH020,abs-1811-09953}. A honest but curious cloud-based machine learning service hosts a network, which is trained on plaintext data (such as public datasets) and hence the network weights are not encrypted \cite{Gilad-BachrachD16}. To ensure the confidentiality of client’s data, the client could encrypt the sensitive segment of data by using HE scheme and send data to the cloud server for performing private inference service without decrypting data or accessing the client’s private key. Only the client could decrypt the inference result by using the private key.

\begin{figure*}[!t]
\centering
\begin{minipage}[t]{0.47\textwidth}
\centering
\includegraphics[width=\linewidth]{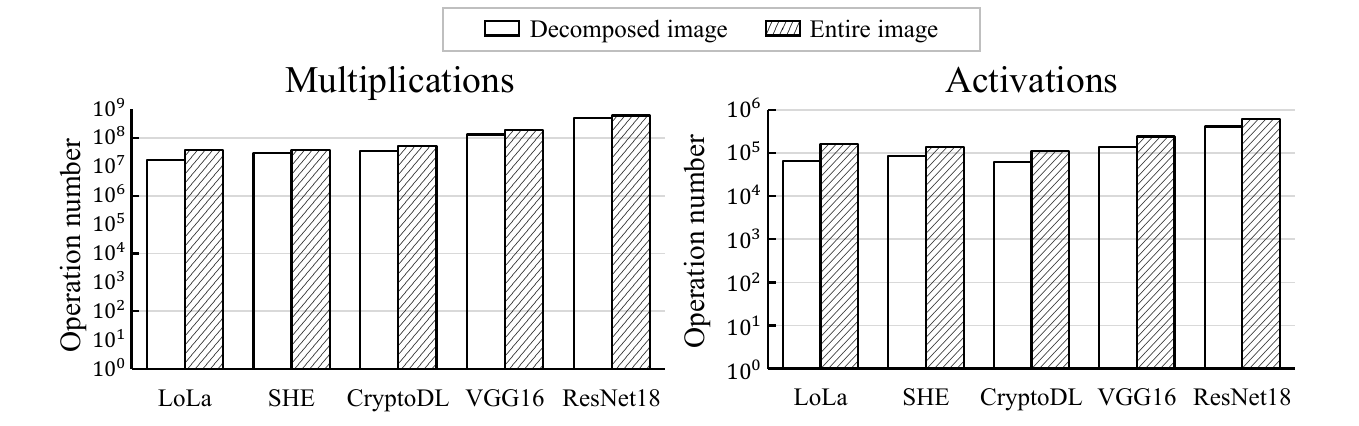}
\caption{Counts of HE multiplications and activations for decomposed and entire image.}
\label{fig:he_seg}
\end{minipage}
\hspace{3mm}
\begin{minipage}[t]{0.47\textwidth}
\centering
\includegraphics[width=\linewidth]{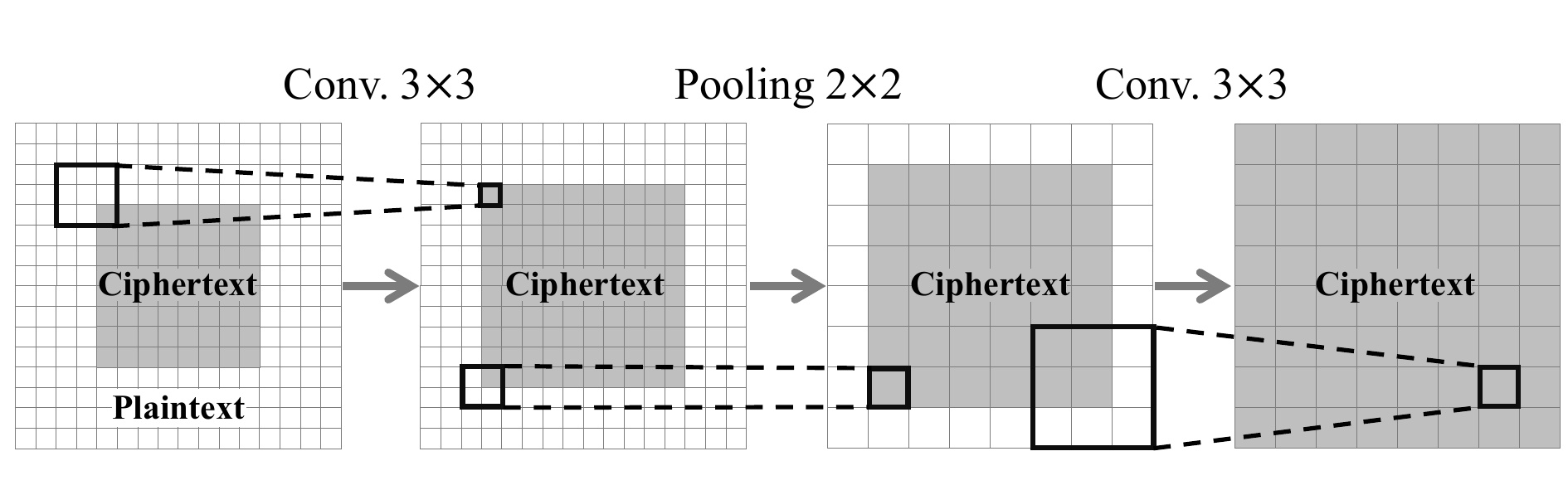}
\caption{The fast spread of ciphertext via CNN layers.}
\label{fig:heops}
\end{minipage}
\vskip -0.25in
\end{figure*}

\section{Our bi-CryptoNets}\label{method}

In this section, we will present new privacy-preserving neural network according to different-level privacy of input data.  Our motivation is to decompose input data into \emph{sensitive} and \emph{insensitive} segments according to their privacy. The sensitive segment includes some important and private information such as human faces in an image, whereas the insensitive one contains some background information, which is not so private yet beneficial to learning algorithm.

Based on such decomposition, we could take the strongest homomorphic encryption to keep data security for sensitive segment, while mingle with some perturbations \cite{dwork2016calibrating,yang2022study} to the data of insensitive segment. This is quite different from previous private inference \cite{Gilad-BachrachD16,LeeLLK0NC22}, where homomorphic encryption is applied to the entire input data.

Notice that we can not directly use some previous neural networks to tackle such decomposition with much smaller computational cost, as shown in Figure~\ref{fig:he_seg}. We compare with three private networks LoLa \cite{BrutzkusGE19}, SHE \cite{Lou019}, CryptoDL \cite{HesamifardTG19}, and two conventional networks VGG16 and ResNet18. Prior networks can not reduce HE operations because ciphertexts could quickly spread over the entire image via several convolution and pooling operations, as shown in Figure~\ref{fig:heops}.

Our idea is quite simple and intuitive for the decomposition of input data. We construct a bi-branch neural network to deal with the sensitive and insensitive segments, which are called \emph{ciphertext} and \emph{plaintext} branch, respectively. The ciphertext branch can make use of features from plaintext branch by unidirectional connections, while the converse direction does not hold because of the quick spread of ciphertexts. We take the feature integration for final predictions from the outputs of two branches of network.

\begin{figure*}[!tb]
\vskip 0.1in
\centering
\includegraphics[width=0.95\linewidth]{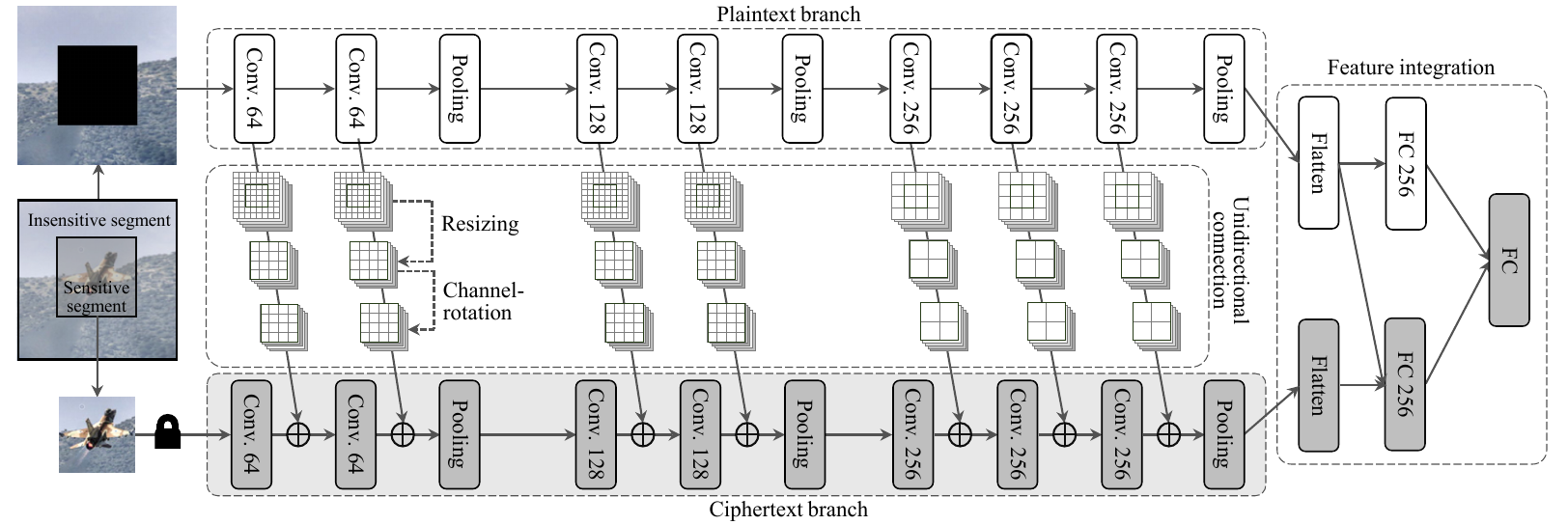}
\caption{The overview of our proposed bi-CryptoNets, where grey layers are computed with encrypted inputs, and other layers are computed with plaintext inputs.}
\label{fig:bi-CryptoNets}
\vskip -0.1in
\end{figure*}

Figure~\ref{fig:bi-CryptoNets} presents an overview for our new privacy-preserving neural network, and it is short for \emph{bi-CryptoNets}. We will go to the details of bi-CryptoNets in the following.

\subsection{The bi-branch of neural network}
We construct ciphertext and plaintext branches to deal with the sensitive and insensitive segments of an input instance, respectively. The plaintext branch deals with the entire input instance, where the insensitive segment is mingled with some perturbations  \cite{dwork2016calibrating,yang2022study}, while the sensitive segment is simply filled with zero. The ciphertext branch tackles the sensitive segment with the homomorphic encryption due to its privacy.

The HE operations are restricted in the ciphertext branch to avoid the ciphertext spread on plaintext branch. This could keep the proportion of HE operations in a stable level, rather than previous close approximation to $1$ as depth increases \cite{Gilad-BachrachD16}. Hence, our bi-branch structure could decease HE operations. Moreover, plaintext branch can be computed with full precision for performing inference, rather than quantized homomorphic ciphertext in prior privacy-preserving neural networks \cite{HesamifardTG19}.

\subsection{The unidirectional connections}\label{sec:uni}

It is well-known that one great success of deep learning lies in the strong features or representations by plentiful co-adaptations of neural network \cite{abs-1207-0580,srivastava2014dropout,goodfellow2016deep}. Our bi-branch neural network reduces some co-adaptations of features obviously, to restrict the spread of ciphertexts. Hence, it is necessary to strengthen features' representations by exploiting some correlations in bi-branches.

Notice that the ciphertext branch can make use of features from plaintext branch, while the converse direction does not hold because of the spread of ciphertexts. In this work, we consider the simple addition to utilize features from plaintext branch as convolutional neural network \cite{resnet16,TanL19,RadosavovicKGHD20}. Concatenation is another effective way to utilize features \cite{IandolaMAHDK16,HuangLMW17}, while it yields more HE operations and computational costs.

\begin{wraptable}{r}{5.7cm}\vskip -0.2in\footnotesize
\caption{The comparisons of latency (ms) for three schemes and four operations.}\label{HEs}\vspace{-5mm}
\begin{center}
\begin{tabular}{|l|cccc|}
\hline
HE Scheme & \text{Add}$_\text{PC}$ & Add$_\text{CC}$ & Mul$_\text{PC}$ & Mul$_\text{CC}$ \\
\hline
BGV & \textbf{0.049} & 0.077 & 2.055 & 3.379\\
CKKS & \textbf{0.039} & 0.077 & 0.173 & 0.390\\
TFHE & \textbf{56.03} & 256.8 & 1018 & 1585\\
\hline
\end{tabular}
\end{center}
\vskip -0.4in
\end{wraptable}

One reason for addition is of small computational costs, since it yields relatively few homomorphic additions between plaintext and ciphertext, as shown in Table~\ref{HEs}. We compare with addition (Add$_\text{PC}$) and multiplication (Mul$_\text{PC}$) between a plaintext and a ciphertext, as well as addition (Add$_\text{CC}$) and multiplication (Mul$_\text{CC}$) between two ciphertexts.  We compare with four operations under three popular HE schemes: BGV \cite{brakerski2014leveled}, CKKS \cite{CheonKKS17} and TFHE \cite{chillotti2020tfhe}, and it is evident that the Add$_\text{PC}$ has the smallest latency.

For unidirectional connections, we first resize the feature map of plaintext branch to fit the addition with ciphertext branch, and then introduce a channel-rotation for ciphertext channel to extract some information from multiple plaintext channels. Specifically, suppose there are $L$ layers in ciphertext and plaintext branches, and denote by $\mathcal{F}_{\text{c},l}$ and $\mathcal{F}_{\text{p},l}$ their respective $l$-th layer.  For every $l\in[L]$, we consider the following two steps:\vspace{+1mm}

\textbf{i) Resizing feature map of plaintext branch}\vspace{+1mm}

We begin with the resizing function
\begin{equation*}
\y_{\text{p},l}=\text{Resize}_{l}(\x_{\text{p},l})\colon \R^{h_{\text{p},l}\times w_{\text{p},l}\times \text{ch}_{\text{p},l}}\to \R^{h_{\text{c},l}\times w_{\text{c},l}\times \text{ch}_{\text{p},l}}
\end{equation*}
where $\x_{\text{p},l}\in\R^{h_{\text{p},l}\times w_{\text{p},l}\times \text{ch}_{\text{p},l}}$ denotes the output feature maps from the $l$-th plaintext layer $\mathcal{F}_{\text{p},l}$ with $h_{\text{p},l}$ rows, $w_{\text{p},l}$ columns and $\text{ch}_{\text{p},l}$ channels, and $h_{\text{c},l}$ and $w_{\text{c},l}$ denote the number of row and column in ciphertext branch.

For resizing function, a simple yet effective choice is the cropping \cite{RonnebergerFB15}, which maintains the center features of size $h_{\text{c},l}\times w_{\text{c},l}\times \text{ch}_{\text{p},l}$ because those features can capture more important information around the sensitive segment. We can also select pooling or convolution operations for resizing function \cite{resnet16}.\vspace{+1mm}

\textbf{ii) Channel-rotation for ciphertext channel}\vspace{+1mm}

We now introduce the channel-rotation function as follows:
\[
\z_{\text{p},l}=\text{CRot}_l(\y_{\text{p},l})=\left[\text{CRot}^1_l(\y_{\text{p},l}),\cdots,\text{CRot}^i_l(\y_{\text{p},l}),\cdots,\text{CRot}^{\text{ch}_{\text{c},l}}_l(\y_{\text{p},l})\right]
\]
where $\text{ch}_{\text{c},l}$ is number of channels in ciphertext branch, and $\text{CRot}^i_l(\y_{\text{p},l})$ denotes the $i$-th channel of $\text{CRot}_l(\x)$, that is,
\[
\text{CRot}^i_l(\y_{\text{p},l})= \sum_{j=1}^{\text{ch}_{\text{p},l}}\W_{\text{CRot},l}^{i,j}\cdot\y_{\text{p},l}^j\ \ \text{ for }\ \  i\in[\text{ch}_{\text{c},l}]\ .
\]
Here, $\y_{\text{p},l}^j\in\R^{h_{\text{c},l}\times w_{\text{c},l}}$ denotes the $j$-th channel of $\y_{\text{p},l}$ for $j\in[\text{ch}_{\text{p},l}]$, and $\W_{\text{CRot},l}=(\W_{\text{CRot},l}^{i,j})_{\text{ch}_{\text{c},l}\times \text{ch}_{\text{p},l}}$ is the weight matrix for channel-rotation in the $l$-th layer.

The channel-rotation is helpful for ciphertext channels to extract useful information automatically from plaintext channels. This is because  channel-rotation could compress, rotate and scale different channels of plaintext feature maps, without the requirements of channel-wise alignments and identical numbers and magnitudes of channels. Moreover, each channel, in connection with ciphertext branch, can be viewed as a linear combination of multiple plaintext channels, rather than one plaintext channel. Therefore, the ciphertext branch can get better information from plaintext branch and achieve better performance with the help of channel-rotation.

\

After resizing and channel-rotation, the ciphertext branch can make use of features from plaintext branch by addition. The unidirectional connections can be written as
\[
\x_{\text{c},l+1}=\mathcal{F}_{\text{c},l+1}\left(\x_{\text{c},l}+\z_{\text{p},l}\right)=\mathcal{F}_{\text{c},l+1}\left(\x_{\text{c},l}+\text{CRot}_l\left(\text{Resize}_l(\x_{\text{p},l})\right)\right)\ ,
\]
where  $\x_{\text{c},l}\in\R^{h_{\text{c},l}\times w_{\text{c},l}\times \text{ch}_{\text{c},l}}$ denotes the output feature maps of $\mathcal{F}_{\text{c},l}$ with $h_{\text{c},l}$ rows, $w_{\text{c},l}$ columns and $\text{ch}_{\text{c},l}$ channels.

Finally, it is feasible to improve computational efficiency by implementing plaintext and ciphertext branches in parallel, because computing the plaintext branch and unidirectional connections is much faster than that of ciphertext branch.

\subsection{The feature integration}

We now design a two-layer neural network to integrate the outputs from ciphertext and plaintext branches. In the first layer, we split all neurons into two halves, one half for ciphertext yet the other for plaintext. The plaintext neurons are only connected with the outputs from plaintext branch, whereas the ciphertext neurons have full connections. We take full connections in the second layer.

Specifically, there are $n_1$ (even) and $n_2$ neurons in the first and second layer, respectively. Let $\x_{\text{c}}\in\R^{n_\text{c}}$ and $\x_{\text{p}}\in\R^{n_\text{p}}$ denote the flattened outputs from ciphertext and plaintext branches with $n_\text{c}$ and $n_\text{p}$ features, respectively. Then, the outputs of ciphertext and plaintext neurons in the first layer can be given by, respectively,
\begin{eqnarray*}
\x_\text{p}^{\langle1\rangle}= \sigma(\W'_{\text{p},1}\x_{\text{p}}+\bm{b}'_1) \ ,\quad
\x_\text{c}^{\langle1\rangle}=\sigma(\W_{\text{c},1}\x_{\text{c}}+\W_{\text{p},1}\x_{\text{p}}+\bm{b}_1)\ ,
\end{eqnarray*}
where $\sigma(\cdot)$ is an activation function, $\bm{b}_1, \bm{b}'_1\in\R^{n_1/2}$ are bias vectors,  $\W_{\text{c},1}\in\R^{n_{\text{c}}\times n_1/2}$ and $\W_{\text{p},1}\in\R^{n_{\text{p}}\times n_1/2}$ are the weight for ciphertext neurons, yet $\W'_{\text{p},1}\in\R^{n_{\text{p}}\times n_1/2}$ is the weight for plaintext neurons. The final output in the second layer can be given by
\[
\x_{\text{out}}=\sigma(\W_{\text{c},2}\x_\text{c}^{\langle1\rangle} +\W_{\text{p},2}\x_\text{p}^{\langle1\rangle}+\bm{b}_2)\ ,
\]
where $\bm{b}_2\in\R^{n_2}$ is a bias vector, and $\W_{\text{c},2}\in\R^{n_1/2\times n_2}$ and $\W_{\text{p},2}\in\R^{n_1/2\times n_2}$ are the weight matrices for ciphertext and plaintext neurons, respectively.
Here, we consider two-layer neural network for simplicity, and similar constructions could be made for deeper neural networks based on the splitting of plaintext and ciphertext neurons.

\section{Knowledge Distillation for bi-CryptoNets}\label{fkd}

\begin{wrapfigure}{r}{6cm}\vskip -0.3in
\centering
\includegraphics[width=6cm]{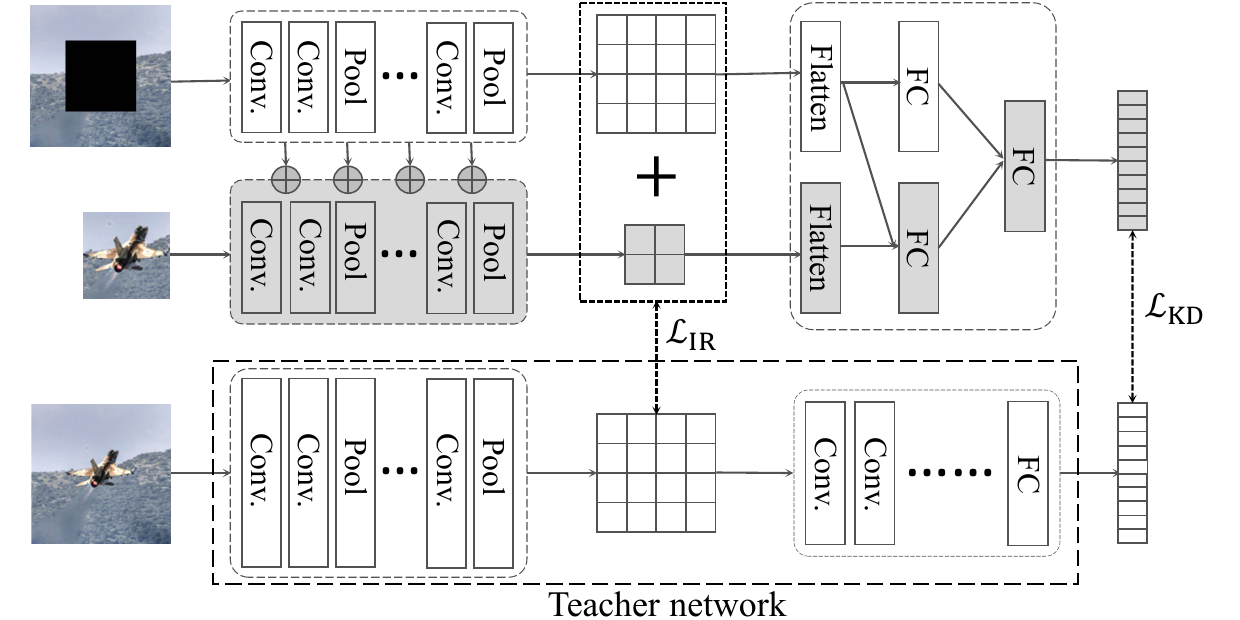}
\caption{The overview of our feature-based knowledge distillation.}
\label{fig:MFKD}
\vskip -0.3in
\end{wrapfigure}

Knowledge distillation has been an effective way to improve learning performance by distilling knowledge from a teacher network \cite{HintonVD15,gou2021knowledge}. This section develops a feature-based knowledge distillation for bi-CryptoNets, as shown in Figure~\ref{fig:MFKD}. The basic idea is to supplement the representations of two branches of bi-CryptoNets by imitating a teacher network.

For the teacher network, we learn a classical convolutional neural network from training data over the entire images without decomposition. Hence, the teacher network has plentiful connections between different neurons, and strengthens the intrinsic correlations for better performance.

We learn the representations of our two branches from the corresponding intermediate representations of teacher network, and also learn the final outputs of our bi-CryptoNets from that of teacher network. This is partially motivated from previous knowledge distillations on internal representations \cite{fitnet}.

Specifically, we first pad the output of ciphertext branch with $0$ to match the size of plaintext output and add them together, as in convolutional neural network literature \cite{goodfellow2016deep,hashemi2019enlarging}. We then train two branches of bi-CryptoNets to learn teacher's intermediate representations by minimizing the following loss function:
\begin{multline}\label{Eqn:knowdis1}
\mathcal{L}_\text{IR}(\W_\text{T}^h,\W_\text{c},\W_\text{p})=\big\|\mathcal{F}_{T}^h(\x,\W_\text{T}^h)
-\left(\text{Pad}(\mathcal{F}_{\text{c}}(\x,\W_\text{c}))+\mathcal{F}_{\text{p}}(\x,\W_\text{p})\right)\big\|^2_2\ ,
\end{multline}
where $\mathcal{F}_{T}^h(\cdot;\W_\text{T}^h)$ is the first $h$ layers of teacher network of parameter $\W_\text{T}^h$, and  $\mathcal{F}_{\text{c}}(\cdot;\W_\text{c})$ and $\mathcal{F}_{\text{p}}(\cdot;\W_\text{p})$ denote the ciphertext and plaintext branches of parameter $\W_\text{c}$ and $\W_\text{p}$, respectively, and $\text{Pad}(\cdot)$ is the zero-padding function.

Here, we denote by $h$ the corresponding $h$-th layer in the teacher network that has the same output size as plaintext branch's output.
Such loss can help the sum of outputs from two branches to approximate the teacher's intermediate output. In this manner, two branches could obtain stronger representations, and this makes it much easier for further learning in the following layers of our bi-CryptoNets.

After training two branches, we then perform knowledge distillation to the whole network. The student network is trained such that its output is similar to that of the teacher and to the true labels. Given bi-CryptoNets' output $\a_S$ and teacher's output $\a_T$, we first get softened output as
\[
\p_T^\tau=\text{softmax}({\a_T}/{\tau})\ \ \text{ and }\ \  \p_S^\tau=\text{softmax}({\a_S}/{\tau})\ ,
\]
where $\tau>1$ is a relaxation parameter. We try to optimize the following loss function:
\begin{equation}\label{Eqn:knowdis2}
\mathcal{L}_\text{KD}(\W_\text{T},\W_\text{BCN})=\mathcal{H}(\y_\text{True},\p_s)+\lambda\mathcal{H}(\p_T^\tau,\p_s^\tau)\ ,
\end{equation}
where $\y_\text{True}$ is the true label, $\p_s=\text{softmax}(\a_S)$, $\mathcal{H}$ refers to the cross-entropy, $\W_\text{T}$ and $\W_\text{BCN}$ are the parameters of teacher network and our bi-CryptoNets, respectively, and $\lambda\in[0,1]$ is a hyperparameter balancing two cross-entropies.
With our proposed method, two branches in our bi-CryptoNets can learn better representations, enhancing overall performance. More details can be found in Appendix~\ref{app:visual}.

\section{Experiments}\label{exp}
This section presents empirical studies on datasets\footnote{Downloaded from \url{yann.lecun.com/exdb/mnist} and \url{www.cs.toronto.edu/~kriz/cifar}} \textsf{MNIST}, \textsf{CIFAR-10} and \textsf{CIFAR-100}, which have been well-studied in previous private inference studies \cite{BrutzkusGE19,Lou019,LeeLLK0NC22}. 
We develop different backbones for our bi-CryptoNets according to different datasets. We adopt the 3-layer CNN \cite{Gilad-BachrachD16,BrutzkusGE19} as backbone for \textsf{MNIST}, and we take 11-layer CNN \cite{ChabanneWMMP17}, VGG-16 \cite{simonyan2014very} and ResNet-18 \cite{resnet16} for \textsf{CIFAR-10} and \textsf{CIFAR-100}. Figure~\ref{fig:bi-CryptoNets} presents our bi-CryptoNets with VGG-16 backbone, and more details are shown in Appendix~\ref{app:net}.

We take the CKKS scheme for sensitive segment with polynomial degree $N = 2^{14}$, and mingle with Gaussian noise for insensitive segment. We further improve packing method for our bi-CryptoNets to perform inference for multiple images simultaneously.
% More details on implementation could be found in Appendix~\ref{app:imp}. 

For simplicity, we focus on the regular images, where the area of sensitive segment is restricted to a quarter in the center of every image, and we could take some techniques of zooming, stretching and rotating to adjust those irregular images.

\

\noindent\textbf{Ablation studies}\vspace{+1mm}

We conduct ablation studies to verify the effectiveness of feature-based knowledge distillation and the structures of our bi-CryptoNets. Table~\ref{table1} presents the details of experimental results of inference accuracies on three datasets.

We first exploit the influence of feature-based knowledge distillation in our bi-CryptoNets. We consider two different methods: bi-CryptoNets without knowledge distillation and bi-CryptoNets with conventional knowledge distillation. As can be seen from Table~\ref{table1}, it is observable that our feature-based knowledge distillation could effectively improve the inference accuracy by $0.54\%\sim 8.39\%$, in comparison to bi-CryptoNets without knowledge distillation; it also achieves better inference accuracy by $1.45\%\sim 10.27\%$ than that of conventional knowledge distillation, which fails to stably improve accuracy with our bi-branch structure.

We then study the influence of unidirectional connections in bi-CryptoNets, and consider two variants: our bi-CryptoNets without unidirectional connections and bi-CryptoNets with resizing yet without channel-rotations. As can be seen from Table~\ref{table1}, the unidirectional connections are helpful for ciphertext branch to extract useful information from plaintext branch. This is because the unidirectional connections with only resizing can enhance inference accuracy by $0.13\%\sim 5.40\%$, and it could further improve accuracy by $0.26\%\sim 5.61\%$ with both resizing and channel-rotations.

We finally compare the inference accuracies of our bi-CryptoNets with the backbone network, which is trained and tested on the entire input images without decomposition. From Table~\ref{table1}, it is observable that our bi-CryptoNets with feature-based knowledge distillation achieves comparable or even better inference accuracies than the corresponding backbone networks without data decomposition; therefore, our proposed bi-CryptoNets could effectively compensate for the information loss of plaintext and ciphertext branches under the help of feature-based knowledge distillation.

\begin{table*}[t]
\caption{The accuracies (\%) on \textsf{MNIST}, \textsf{CIFAR-10} and \textsf{CIFAR-100} datasets (KD refers to the conventional knowledge distillation; FKD refers to our feature-based knowledge distillation).}\label{table1}\vspace{-2mm}
\begin{center}
\begin{small}
\begin{tabular}{|l|c|c|ccc|cc|}
\toprule
\multirow{2}{*}{ \scriptsize Models} & \scriptsize  Knowledge  & \scriptsize  \textsf{MNIST}  & \multicolumn{3}{c|}{\scriptsize\textsf{CIFAR-10}} & \multicolumn{2}{c|}{\scriptsize\textsf{CIFAR-100}} \\
& \scriptsize  Distillation & \scriptsize  CNN-3 & \scriptsize  CNN-11 & \scriptsize  VGG-16 & \scriptsize  ResNet-18 & \scriptsize  VGG-16 & \scriptsize  ResNet-18\\
\midrule
\scriptsize Backbone network (w/o decomposition) & \scriptsize  w/o KD & \scriptsize  $99.21$ & \scriptsize  $90.99$ & \scriptsize  $93.42$& \scriptsize  $94.30$ & \scriptsize $72.23$ & \scriptsize  $74.00$\\
\midrule
\scriptsize {bi-CryptoNets } & \scriptsize  w/o KD & \scriptsize  $98.60$ & \scriptsize  $81.91$ & \scriptsize  $90.36$& \scriptsize  $92.04$ & \scriptsize  $64.85$& \scriptsize  $69.46$ \\
\scriptsize {bi-CryptoNets }& \scriptsize  KD & \scriptsize  $98.61$ & \scriptsize  $80.03$ & \scriptsize  $88.91$& \scriptsize  $92.43$ & \scriptsize  $64.96$& \scriptsize  $65.67$ \\
\scriptsize bi-CryptoNets (w/o uni. connections) & \scriptsize  FKD & \scriptsize  $98.89$ & \scriptsize  $84.69$ & \scriptsize  $91.78$ & \scriptsize  $91.73$ & \scriptsize  $70.61$ & \scriptsize  $70.29$\\
\scriptsize bi-CryptoNets (w/o channel-rotations)& \scriptsize  FKD & \scriptsize  $98.90$ & \scriptsize  $90.09$ & \scriptsize  $92.14$& \scriptsize  $91.86$ & \scriptsize  $71.43$& \scriptsize  $71.40$ \\
\scriptsize \textbf{bi-CryptoNets}& \scriptsize  \textbf{FKD} & \scriptsize  $\mathbf{99.15}$ & \scriptsize $\mathbf{90.30}$ & \scriptsize  $\mathbf{93.27}$& \scriptsize  $\mathbf{93.91}$ & \scriptsize  $\mathbf{72.35}$& \scriptsize  $\mathbf{73.33}$ \\
\bottomrule
\end{tabular}
\end{small}
\end{center}\vspace{-8mm}
\end{table*}

\ 

\noindent\textbf{Experimental comparisons}\vspace{+1mm}

We compare our bi-CryptoNets with the state-of-the-art schemes on privacy-preserving neural networks \cite{abs-1811-09953,HesamifardTG19,BrutzkusGE19,Lou019,DathathriKSDLM20,LeeLLK0NC22}. We also implement the structure of backbone network with square activation for fair comparisons. We take the batch size $20$ and $32$ for  \textsf{MNIST} and \textsf{CIFAR-10}, respectively.
We employ commonly-used criteria in experiments, i.e., inference accuracy, inference latency and the number of homomorphic operations as in  \cite{abs-1811-09953,LouLH020}. We also consider the important amortized inference latency in a batch of images \cite{LeeLLK0NC22}, and concern the number of activations for ciphertexts \cite{Lou019,BrutzkusGE19,DathathriKSDLM20}. Tables~\ref{table2} and \ref{table3} summarize experimental comparisons on \textsf{MNIST} and \textsf{CIFAR-10}, respectively. Similar results could be made for \textsf{CIFAR-100}, presented in Appendix~\ref{app:100}.

From Table~\ref{table2}, our bi-CryptoNets can reduce HE operations by $2.35\times$ and inference latency by $3.43\times$ in contrast to backbone network. Our bi-CryptoNets could decrease the amortized latency by $13.7\times$, and achieve better accuracy (about $0.2\%$) than backbone because of our knowledge distillation and precise inference in the plaintext branch.

From Table~\ref{table3}, our bi-CryptoNets reduces HE operations by $1.43\times$, and inference latency by $1.15\times$ in contrast to backbone network. Our bi-CryptoNets could decrease the amortized latency by $4.59\times$ and improve accuracy by $2.1\%$. We also implement our bi-CryptoNets with VGG-16 and ResNet-18 as backbones, and deeper networks yield better inference accuracies, i.e., $93.27\%$ for VGG-16 and $93.91\%$ for ResNet-18.

From Tables~\ref{table2} and \ref{table3}, our bi-CryptoNets takes a good balance between inference accuracies and computation cost in comparison with other neural networks. Our bi-CryptoNets achieves lower inference latency and amortized latency than other methods except for LoLa, where light-weight neural network is implemented with complicated packing method and smaller HE operations. Our bi-CryptoNets takes relatively-good inference accuracies except for SHE, CryptoDL and VDSCNN, where deeper neural networks are adopted with larger computational overhead. Based on deeper backbones such as VGG-16 and ResNet-18, our bi-CryptoNets could achieve better inference accuracy, comparable inference latency and smaller amortized latency.

\begin{table*}[t]\scriptsize
\caption{The experimental comparisons on \textsf{MNIST}.}\label{table2}\vspace{-2mm}
\begin{center}
\begin{small}
\begin{tabular}{lcccccccccc}
\toprule
\scriptsize Scheme & \scriptsize HEOPs & \scriptsize Add$_\text{CC}$ & \scriptsize Mul$_\text{PC}$ & \scriptsize Act$_\text{C}$ & \scriptsize Latency(s) & \scriptsize Amortized Latency(s) & \scriptsize Acc(\%)\\
\midrule
\scriptsize {FCryptoNets \cite{abs-1811-09953}} & \scriptsize  63K & \scriptsize  38K & \scriptsize  24K & \scriptsize  945 & \scriptsize  39.1 & \scriptsize  2.0 & \scriptsize  98.71\\
\scriptsize {CryptoDL \cite{HesamifardTG19}} & \scriptsize  4.7M & \scriptsize  2.3M & \scriptsize  2.3M & \scriptsize  1.6K & \scriptsize  320 & \scriptsize  16.0 & \scriptsize  99.52\\
\scriptsize {LoLa \cite{BrutzkusGE19}} & \scriptsize  573 & \scriptsize  393 & \scriptsize  178 & \scriptsize  2 & \scriptsize  2.2 & \scriptsize  2.2 & \scriptsize  98.95 \\
\scriptsize {SHE \cite{Lou019}} & \scriptsize  23K & \scriptsize  19K & \scriptsize  945 & \scriptsize  3K & \scriptsize  9.3 & \scriptsize  9.3 & \scriptsize  99.54\\
% {DSHE \cite{Lou019}} & \scriptsize  314K & \scriptsize  304K & \scriptsize  5K & \scriptsize  5K & \scriptsize  124.9 & \scriptsize  124.9 & \scriptsize  99.77\\
\scriptsize {EVA \cite{DathathriKSDLM20}} & \scriptsize  8K & \scriptsize  4K & \scriptsize  4K & \scriptsize  3 & \scriptsize  121.5 & \scriptsize  121.5 & \scriptsize  99.05\\
\scriptsize {VDSCNN \cite{LeeLLK0NC22}} & \scriptsize  4K & \scriptsize  2K & \scriptsize  2K & \scriptsize  48 & \scriptsize  105 & \scriptsize  105 & \scriptsize  99.19\\
\scriptsize {Backbone CNN-3} & \scriptsize  1962 & \scriptsize  973 & \scriptsize  984 & \scriptsize  5 & \scriptsize  7.2 & \scriptsize  1.4 & \scriptsize  98.95\\
\midrule
\scriptsize {\textbf{bi-CryptoNets (CNN-3)}} & \scriptsize  830 & \scriptsize  406 & \scriptsize  418 & \scriptsize  6 & \scriptsize  2.1 & \scriptsize  0.1 & \scriptsize  99.15 \\
\bottomrule
\end{tabular}
\end{small}
\end{center}\vspace{-8mm}
\end{table*}

\begin{table*}[t]
\caption{The experimental comparisons on \textsf{CIFAR-10}.}\label{table3}\vspace{-2mm}
\begin{center}
\begin{small}
\begin{tabular}{lcccccccccc}
\toprule
\scriptsize Scheme & \scriptsize HEOPs & \scriptsize Add$_\text{CC}$ & \scriptsize Mul$_\text{PC}$ & \scriptsize Act$_\text{C}$ & \scriptsize Latency(s) & \scriptsize Amortized Latency(s) & \scriptsize Acc(\%)\\
\midrule
\scriptsize{FCryptoNets \cite{abs-1811-09953}} & \scriptsize 701M & \scriptsize 350M & \scriptsize 350M & \scriptsize 64K & \scriptsize 22372 & \scriptsize 1398 & \scriptsize 76.72\\
\scriptsize{CryptoDL \cite{HesamifardTG19}} & \scriptsize 2.4G & \scriptsize 1.2G & \scriptsize 1.2G & \scriptsize 212K & \scriptsize 11686 & \scriptsize 731 & \scriptsize 91.50\\
\scriptsize {LoLa \cite{BrutzkusGE19}} & \scriptsize 70K & \scriptsize 61K & \scriptsize 9K & \scriptsize 2 & \scriptsize 730 & \scriptsize 730 & \scriptsize 76.50\\
\scriptsize {SHE \cite{Lou019}} & \scriptsize 4.4M & \scriptsize 4.4M & \scriptsize 13K & \scriptsize 16K & \scriptsize 2258 & \scriptsize 2258 & \scriptsize 92.54\\
% {DSHE \cite{Lou019}} & 68M & 68M & 98K & 131K & 12041 & 12041 & 94.62\\
\scriptsize{EVA \cite{DathathriKSDLM20}} & \scriptsize 135K & \scriptsize 67K & \scriptsize 67K & \scriptsize 9 & \scriptsize 3062 & \scriptsize 3062 & \scriptsize 81.50\\
\scriptsize{VDSCNN \cite{LeeLLK0NC22}} & \scriptsize 18K & \scriptsize 8K & \scriptsize 9K & \scriptsize 752 & \scriptsize 2271 & \scriptsize 2271 & \scriptsize 91.31\\
\scriptsize{Backbone CNN-11} & \scriptsize 1.0M & \scriptsize 493K & \scriptsize 521K & \scriptsize 246 & \scriptsize 1823 & \scriptsize 228 & \scriptsize 88.21\\
\midrule
\scriptsize {\textbf{bi-CryptoNets (CNN-11)}} & \scriptsize 709K & \scriptsize 341K & \scriptsize 368K & \scriptsize 246 & \scriptsize 1587 & \scriptsize 49 & \scriptsize 90.30\\
\scriptsize {\textbf{bi-CryptoNets (VGG-16)}} & \scriptsize 3.4M & \scriptsize 1.5M & \scriptsize 1.9M & \scriptsize 1.1K & \scriptsize 2962 & \scriptsize 92 & \scriptsize 93.27\\
\scriptsize {\textbf{bi-CryptoNets (ResNet-18)}} & \scriptsize 15.5M & \scriptsize 6.9M & \scriptsize 8.6M & \scriptsize 2.8K & \scriptsize 6760 & \scriptsize 211 & \scriptsize 93.91\\
\bottomrule
\end{tabular}
\end{small}
\end{center}\vspace{-8mm}
\end{table*}

\section{Conclusion}\label{conclusion}
Numerous privacy-preserving neural networks have been developed to keep the balance between accuracy, efficiency and security. We take a different view from the input data and network structure. We decompose the input data into sensitive and insensitive segments, and propose the bi-CryptoNets, i.e., plaintext and ciphertext branches, to deal with two segments, respectively. We also introduce feature-based knowledge distillation to strengthen the representations of our network. Empirical studies verify the effectiveness of our bi-CryptoNets. An interesting future work is to exploit multiple levels of privacy of input data and multiple branches of neural network, and it is also interesting to generalize our idea to other settings such as multi-party computation.

\section*{Acknowledgements}
The authors want to thank the reviewers for their helpful comments and suggestions. This research was supported by National Key R\&D Program of China (2021ZD0112802), NSFC (62376119) and CAAI-Huawei MindSpore Open Fund.

\bibliographystyle{splncs04}
\bibliography{reference}

\begin{thebibliography}{10}
\providecommand{\url}[1]{\texttt{#1}}
\providecommand{\urlprefix}{URL }
\providecommand{\doi}[1]{https://doi.org/#1}

\bibitem{BoemerCCW19}
Boemer, F., Costache, A., Cammarota, R., Wierzynski, C.: {nGraph-HE2}: {A}
  high-throughput framework for neural network inference on encrypted data. In:
  WAHC@CCS. pp. 45--56 (2019)

\bibitem{BoulemtafesDC20}
Boulemtafes, A., Derhab, A., Challal, Y.: A review of privacy-preserving
  techniques for deep learning. Neurocomputing  \textbf{384},  21--45 (2020)

\bibitem{brakerski2014leveled}
Brakerski, Z., Gentry, C., Vaikuntanathan, V.: (leveled) fully homomorphic
  encryption without bootstrapping. {ACM} Trans. Comput. Theory  \textbf{6}(3),
   1--36 (2014)

\bibitem{BrutzkusGE19}
Brutzkus, A., Gilad{-}Bachrach, R., Elisha, O.: Low latency privacy preserving
  inference. In: ICML. pp. 812--821 (2019)

\bibitem{ChabanneWMMP17}
Chabanne, H., Wargny, A., Milgram, J., Morel, C., Prouff, E.:
  Privacy-preserving classification on deep neural network. Cryptol. ePrint
  Arch.  (2017)

\bibitem{CheonKKS17}
Cheon, J., Kim, A., Kim, M., Song, Y.: Homomorphic encryption for arithmetic of
  approximate numbers. In: ASIACRYPT. pp. 409--437 (2017)

\bibitem{chillotti2020tfhe}
Chillotti, I., Gama, N., Georgieva, M., Izabach{\`e}ne, M.: {TFHE}: {Fast}
  fully homomorphic encryption over the torus. J. Cryptol.  \textbf{33}(1),
  34--91 (2020)

\bibitem{abs-1811-09953}
Chou, E., Beal, J., Levy, D., Yeung, S., Haque, A., Fei{-}Fei, L.: Faster
  {CryptoNets}: {Leveraging} sparsity for real-world encrypted inference.
  CoRR/abstract  \textbf{1811.09953} (2018)

\bibitem{DathathriKSDLM20}
Dathathri, R., Kostova, B., Saarikivi, O., Dai, W., Laine, K., Musuvathi, M.:
  {EVA:} {An} encrypted vector arithmetic language and compiler for efficient
  homomorphic computation. In: PLDI. pp. 546--561 (2020)

\bibitem{DathathriS0LLMM19}
Dathathri, R., Saarikivi, O., Chen, H., Laine, K., Lauter, K., Maleki, S.,
  Musuvathi, M., Mytkowicz, T.: {CHET:} {An} optimizing compiler for
  fully-homomorphic neural-network inferencing. In: PLDI. pp. 142--156 (2019)

\bibitem{dwork2016calibrating}
Dwork, C., McSherry, F., Nissim, K., Smith, A.: Calibrating noise to
  sensitivity in private data analysis. J. Priv. Confidentiality
  \textbf{7}(3),  17--51 (2016)

\bibitem{Gentry09}
Gentry, C.: Fully homomorphic encryption using ideal lattices. In: STOC. pp.
  169--178 (2009)

\bibitem{GhodsiJRG21}
Ghodsi, Z., Jha, N., Reagen, B., Garg, S.: Circa: {Stochastic} {ReLUs} for
  private deep learning. In: NeurIPS. pp. 2241--2252 (2021)

\bibitem{Gilad-BachrachD16}
Gilad{-}Bachrach, R., Dowlin, N., Laine, K., Lauter, K., Naehrig, M., Wernsing,
  J.: Cryptonets: Applying neural networks to encrypted data with high
  throughput and accuracy. In: ICML. pp. 201--210 (2016)

\bibitem{GongCYMW21}
Gong, X., Chen, Y., Yang, W., Mei, G., Wang, Q.: {InverseNet}: {A}ugmenting
  model extraction attacks with training data inversion. In: IJCAI. pp.
  2439--2447 (2021)

\bibitem{goodfellow2016deep}
Goodfellow, I., Bengio, Y., Courville, A.: Deep Learning. MIT Press (2016)

\bibitem{gou2021knowledge}
Gou, J., Yu, B., Maybank, S., Tao, D.: Knowledge distillation: {A} survey. Int.
  J. Comput. Vis.  \textbf{129}(6),  1789--1819 (2021)

\bibitem{hashemi2019enlarging}
Hashemi, M.: Enlarging smaller images before inputting into convolutional
  neural network: {Zero-padding} vs. interpolation. J. Big Data  \textbf{6}(1),
   1--13 (2019)

\bibitem{resnet16}
He, K., Zhang, X., Ren, S., Sun, J.: Deep residual learning for image
  recognition. In: CVPR. pp. 770--778 (2016)

\bibitem{HesamifardTG19}
Hesamifard, E., Takabi, H., Ghasemi, M.: Deep neural networks classification
  over encrypted data. In: CODASPY. pp. 97--108 (2019)

\bibitem{abs-1207-0580}
Hinton, G., Srivastava, N., Krizhevsky, A., Sutskever, I., Salakhutdinov, R.:
  Improving neural networks by preventing co-adaptation of feature detectors.
  CoRR/abstract  \textbf{1207.0580} (2012)

\bibitem{HintonVD15}
Hinton, G., Vinyals, O., Dean, J.: Distilling the knowledge in a neural
  network. CoRR/abstract  \textbf{1503.02531} (2015)

\bibitem{HuangLMW17}
Huang, G., Liu, Z., Maaten, L., Weinberger, K.: Densely connected convolutional
  networks. In: CVPR. pp. 2261--2269 (2017)

\bibitem{IandolaMAHDK16}
Iandola, F., Moskewicz, M., Ashraf, K., Han, S., Dally, W., Keutzer, K.:
  {SqueezeNet}: {AlexNet-level} accuracy with 50x fewer parameters and
  {\textless}{1MB} model size. CoRR/abstract  \textbf{1602.07360} (2016)

\bibitem{LeeLLK0NC22}
Lee, E., Lee, J., Lee, J., Kim, Y., Kim, Y., No, J., Choi, W.: Low-complexity
  deep convolutional neural networks on fully homomorphic encryption using
  multiplexed parallel convolutions. In: ICML. pp. 12403--12422 (2022)

\bibitem{LiLYZSLLY23}
Li, Z., Li, X., Yang, L., Zhao, B., Song, R., Luo, L., Li, J., Yang, J.:
  Curriculum temperature for knowledge distillation. In: AAAI. pp. 1504--1512
  (2023)

\bibitem{Lou019}
Lou, Q., Jiang, L.: {SHE:} {A} fast and accurate deep neural network for
  encrypted data. In: NeurIPS. pp. 10035--10043 (2019)

\bibitem{LouLH020}
Lou, Q., Lu, W., Hong, C., Jiang, L.: Falcon: {Fast} spectral inference on
  encrypted data. In: NeurIPS. pp. 2364--2374 (2020)

\bibitem{RadosavovicKGHD20}
Radosavovic, I., Kosaraju, R., Girshick, R., He, K., Doll{\'{a}}r, P.:
  Designing network design spaces. In: CVPR. pp. 10425--10433 (2020)

\bibitem{RibeiroGC15}
Ribeiro, M., Grolinger, K., Capretz, M.: {MLaaS}: {M}achine learning as a
  service. In: ICMLA. pp. 896--902 (2015)

\bibitem{fitnet}
Romero, A., Ballas, N., Kahou, S., Chassang, A., Gatta, C., Bengio, Y.:
  Fitnets: {Hints} for thin deep nets. In: ICLR (2015)

\bibitem{RonnebergerFB15}
Ronneberger, O., Fischer, P., Brox, T.: {U-Net}: {Convolutional} networks for
  biomedical image segmentation. In: MICCAI. pp. 234--241 (2015)

\bibitem{sealcrypto}
{M}icrosoft {SEAL} (release 4.0). \url{https://github.com/Microsoft/SEAL} (Mar
  2022), microsoft Research, Redmond, WA.

\bibitem{simonyan2014very}
Simonyan, K., Zisserman, A.: Very deep convolutional networks for large-scale
  image recognition. CoRR/abstract  \textbf{1409.1556} (2014)

\bibitem{srivastava2014dropout}
Srivastava, N., Hinton, G., Krizhevsky, A., Sutskever, I., Salakhutdinov, R.:
  Dropout: {A} simple way to prevent neural networks from overfitting. JMLR
  \textbf{15}(1),  1929--1958 (2014)

\bibitem{TanL19}
Tan, M., Le, Q.: {EfficientNet}: {Rethinking} model scaling for convolutional
  neural networks. In: ICML. pp. 6105--6114 (2019)

\bibitem{xie2023gini}
Xie, X.R., Yuan, M.J., Bai, X.T., Gao, W., Zhou, Z.H.: On the {G}ini-impurity
  preservation for privacy random forests. In: NeurIPS (2023)

\bibitem{yang2022study}
Yang, K., Yau, J., Fei-Fei, L., Deng, J., Russakovsky, O.: A study of face
  obfuscation in imagenet. In: ICML. pp. 25313--25330 (2022)

\bibitem{YinZH21}
Yin, X., Zhu, Y., Hu, J.: A comprehensive survey of privacy-preserving
  federated learning: {A} taxonomy, review, and future directions. {ACM}
  Comput. Surv.  \textbf{54}(6),  131:1--131:36 (2021)

\bibitem{ZagoruykoK17}
Zagoruyko, S., Komodakis, N.: Paying more attention to attention: {I}mproving
  the performance of convolutional neural networks via attention transfer. In:
  {ICLR} (2017)

\end{thebibliography}

\newpage
\appendix
\onecolumn
\begin{center}
{\Large\textbf{Supplementary Material (Appendix)}}
\end{center}
\section{CKKS Scheme}
In this section, we will describe CKKS scheme in more details. CKKS is a HE scheme that supports fixed-point arithmetic operations on encrypted data.
The ciphertext of the CKKS scheme is an element $\c\in\mathcal{R}^2$, where $\mathcal{R}$ denotes the polynomial ring $\mathcal{R}_p^N=\mathbb{Z}_p[x]/(x^N+1)$, $\mathbb{Z}_p[x]=\mathbb{Z}/p\mathbb{Z}$, and $p=\prod_{i=1}^l q_i$ is a product of $l$ prime numbers, $N$ is the polynomial degree. In CKKS scheme, we can encrypt $n=N/2$ plaintext in $N/2$ slots of a single ciphertext, and perform Single Instruction Multiple Data (SIMD) operations at no extra cost.
Let $E(\cdot)$ and $D(\cdot)$ denote the encryption and decryption function respectively. Let $\c, \c_1, \c_2, \c_\text{add}, \c_\text{mult}, \c_\text{rot}$ denote ciphertext, and $\bm{u}$ denotes a plaintext vector in $\R^n$. CKKS scheme supports following homomorphic operations:
\begin{itemize}
    \item Homomorphic addition:
    \[
        \text{Add}(\c,\bm{u})\to \c_\text{add}\ ,\ \text{ s.t. }\ D(\c_\text{add})=D(\c_1)+\bm{u}\ ,
    \]
    \[
        \text{Add}(\c_1,\c_2)\to \c_\text{add}\ ,\ \text{ s.t. }\ D(\c_\text{add})=D(\c_1)+D(\c_2)\ .
    \]
    \item Homomorphic multiplication:
    \[
        \text{Mult}(\c,\bm{u})\to \c_\text{mult}\ ,\ \text{ s.t. }\ D(\c_\text{mult})=D(\c_1)\cdot\bm{u}\ ,
    \]
    \[
        \text{Mult}(\c_1,\c_2)\to \c_\text{mult}\ ,\ \text{ s.t. }\ D(\c_\text{mult})=D(\c_1)\cdot D(\c_2)\ ,
    \]
    where $\cdot$ is the element-wize multiplication between two vectors.
    \item Homomorphic rotation:
    \[
        \text{Rot}(\c,r)\to \c_\text{rot}\ ,\ \text{ s.t. }\ D(\c_\text{rot})=\langle D(\c)\rangle_r\ ,
    \]
    where $r\in\mathbb Z$, $\langle\cdot\rangle_r$ denotes cyclically shifting a vector by $r$ to the left when $r>0$, and shifting by $-r$ to the right when $r<0$.
\end{itemize}

Gilad-Bachrach et al. performed single instruction multiple data operations without additional costs by packing multiple plaintext into one ciphertext for HE \cite{Gilad-BachrachD16}. Specifically, a vector of $N/2$ fixed-point numbers can be packed into $N/2$ slots of a single ciphertext for CKKS \cite{CheonKKS17}, and element-wize multiplication and addition between two vectors can be performed with only one HE operation. Such technique has been used to reduce HE computations in privacy-preserving neural networks, such as packing multiple channels of input data or multiple neurons into one ciphertext \cite{BrutzkusGE19,DathathriKSDLM20}.

\section{Experiment Details}
\paragraph{Datasets.}
We conduct experiments on \textsf{MNIST}, \textsf{CIFAR-10} and \textsf{CIFAR-100} datasets.
\textsf{MNIST} has $10$ output classes with $60,000$ training images and $10,000$ testing images, and each image's size is $28\times28\times1$. \textsf{CIFAR-10} and \textsf{CIFAR-100} have $10$ and $100$ output classes respectively, and both of them consist of $50,000$ training images and $10,000$ testing images, and each image's size is $32\times32\times3$.

\paragraph{Parameter settings.}
Datasets have been preprocessed by image centering, that is, subtracting mean and dividing standard deviation, and images have been augmented by using the random horizontal flips and random crops. We also adopt the squared activation in ciphertext branch and feature integration, since CKKS only supports homomorphic addition and multiplication. In training, we use Adam optimizer to train our network with batch size of $256$. We adopt the learning rate of $0.0001$ with cosine decay. For feature-based knowledge distillation, the teacher network takes backbone network with ReLU activation and hyper-parameters $\tau=4$ and $\lambda=0.9$. Our experiments are conducted on the HE scheme library SEAL \cite{sealcrypto} on AMD Ryzen Threadripper 3970X at 2.2 GHz with 256GB of RAM, running the Ubuntu 20.04 operating system.

\paragraph{State-of-the-art privacy-preserving neural networks.}

\begin{itemize}
    % \item CryptoNets \cite{Gilad-BachrachD16}: A privacy-preserving neural network using YASHE' scheme to encrypt inputs. All network weights are quantized into integers, and activations are replaced with square activation.
    \item FCryptoNets \cite{abs-1811-09953}: A privacy-preserving neural network that leverages sparse representations in the underlying cryptosystem to accelerate inference. It develops a pruning and quantization approach that achieves maximally-sparse encodings and minimizes approximation error. Such approach can reduce the number of HE operations, and achieve short inference latency.
    \item CryptoDL \cite{HesamifardTG19}:  A privacy-preserving neural network with polynomial activations. It use low degree polynomials to approximate ReLU activations, and implement deeper networks, achieving better performance in accuracy.
    \item LoLa \cite{BrutzkusGE19}: A privacy-preserving neural network with well-designed packing methods. It can pack one image into several ciphertext, and achieve fast inference for one single image with much fewer HE operations. On the other hand, it can only perform inference for one image every time.
    \item SHE \cite{Lou019}: A privacy-preserving neural network implemented with TFHE scheme. It use boolean circuit to implement neural network by quantizing network weights into powers of $2$, and it can support ReLU activation and max pooling. Therefore, it can reach higher accuracy.
    \item EVA \cite{DathathriKSDLM20}: A privacy-preserving neural network that implemented by a compiler for HE with CKKS scheme. The compiler can optimize the hyperparameters and the computation graph for HE. Therefore, it can decrease the HE computation depth and shorten the inference latency for private-preserving neural networks.
    \item VDSCNN \cite{LeeLLK0NC22}: A privacy-preserving neural network implemented with CKKS scheme. It uses polynomial activations and it also packs multiple channels of the image into one ciphertext. Therefore, it can perform inference with fewer HE additions and multiplications.
\end{itemize}

\begin{figure}[!tb]
\vskip 0.1in
\centering
\includegraphics[width=\linewidth]{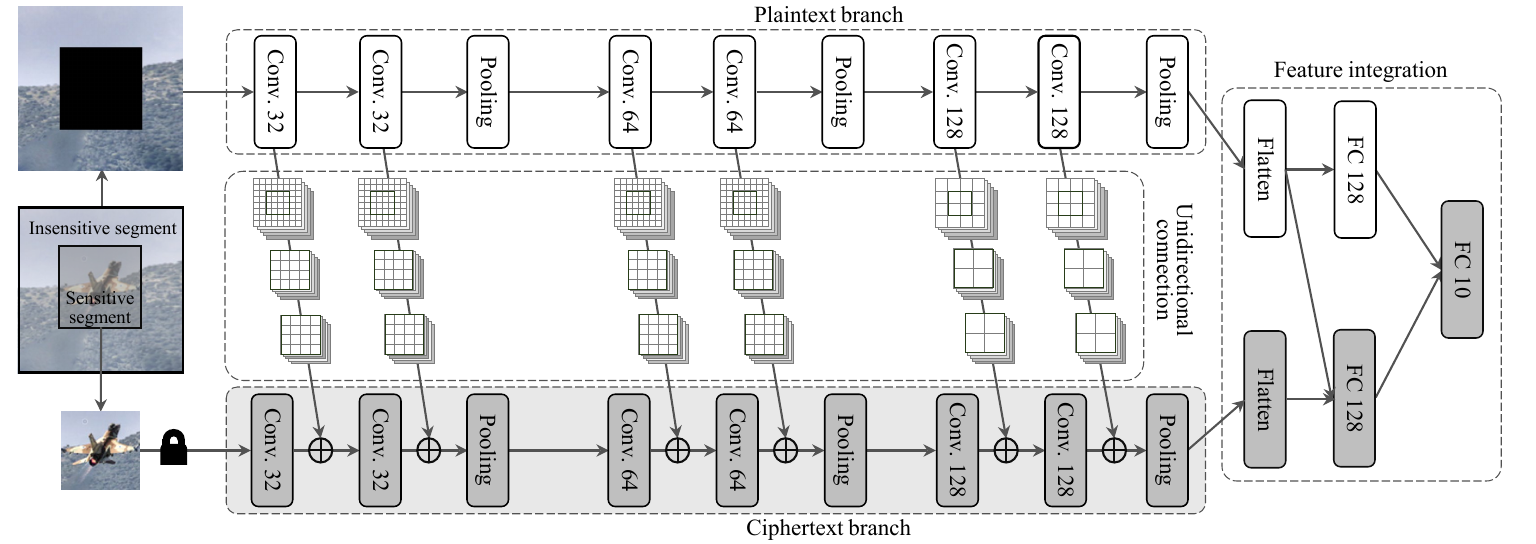}
\caption{The structure of our proposed bi-CryptoNets taking 11-layer CNN as backbone.}
\label{fig:mcn1}
\vskip -0.1in
\end{figure}

\begin{figure}[!tb]
\vskip 0.1in
\centering
\includegraphics[width=\linewidth]{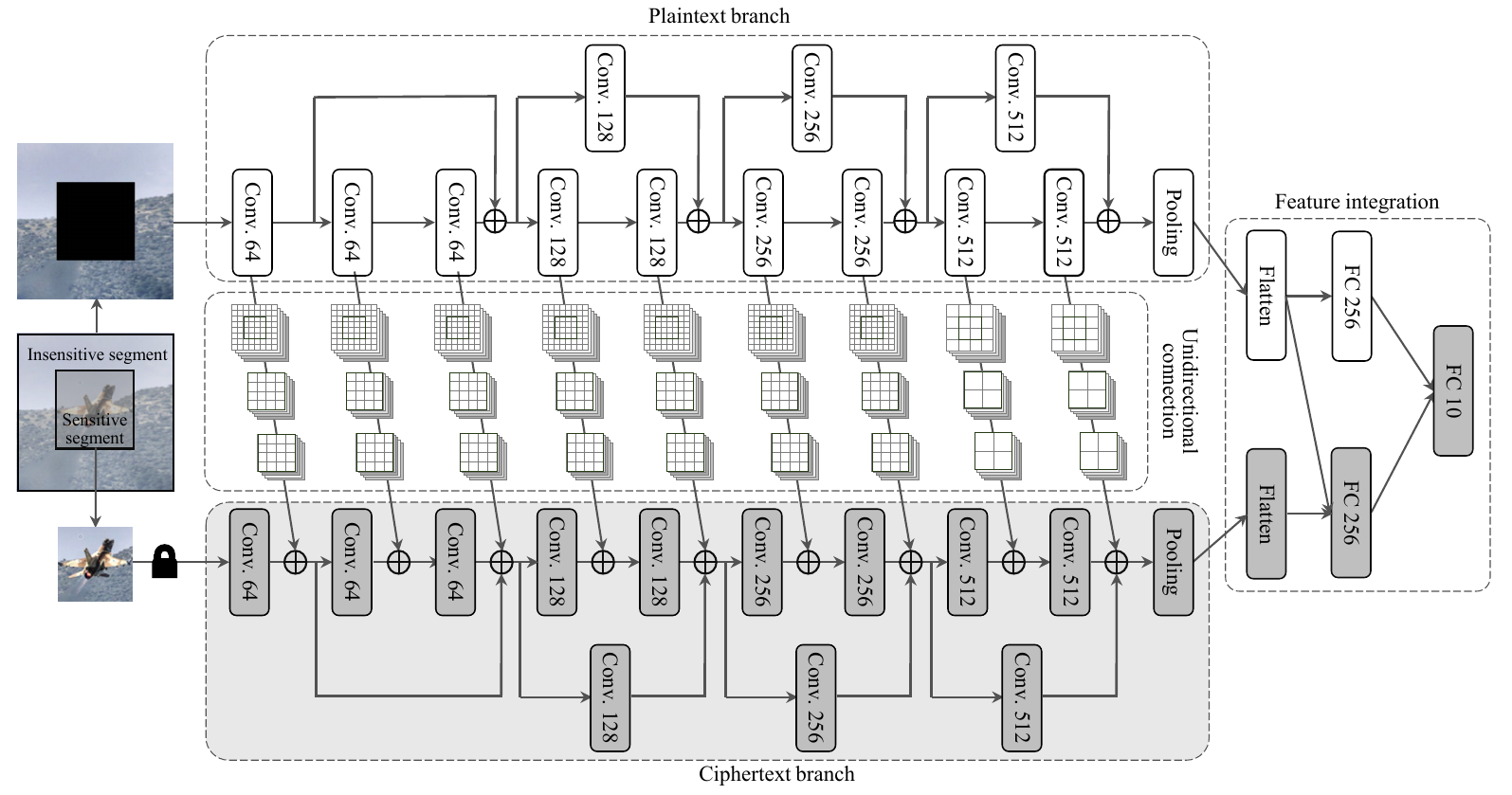}
\caption{The structure of our proposed bi-CryptoNets taking ResNet-18 as backbone.}
\label{fig:mcn2}
\vskip -0.1in
\end{figure}

\section{Network Architectures}
\label{app:net}
In this section, we will present the structure for other variants of our bi-CryptoNets in details.

Figure~\ref{fig:mcn1} presents the overview structure of our proposed bi-CryptoNets that takes 11-layer CNN as backbone. Each branch has the following layout: (i) 2 layers of $3\times 3$ convolution with stride of $(1,1)$, padding of $(1,1)$ and 32 output maps; (ii) $2\times 2$ average pooling with $(2, 2)$ stride; (iii) 2 layers of $3\times 3$ convolution with stride of $(1,1)$, padding of $(1,1)$ and 64 output maps; (iv) $2\times 2$ average pooling with $(2, 2)$ stride; (v) 2 layers of $3\times 3$ convolution with stride of $(1,1)$, padding of $(1,1)$ and 128 output maps; (vi) $2\times 2$ average pooling with $(2, 2)$ stride.
We build unidirectional connections after each convolutional layers in the plaintext branch.
Following two branches, we have a 2-layer feature integration: In the first layer, we have two fully connected layers with 128 plaintext outputs and 128 ciphertext outputs, respectively; in the second layer, we have fully connected layer with 10 ciphertext outputs.

Figure~\ref{fig:mcn2} presents the overview structure of our proposed bi-CryptoNets that takes ResNet-18 as backbone. Each branch has the following layout: (i) $3\times 3$ convolution with stride of $(1,1)$, padding of $(1,1)$ and 64 output maps; (ii) 2 layers of $3\times 3$ convolution with stride of $(1,1)$, padding of $(1,1)$ and 64 output maps; (iii) skip connection by adding the outputs from (i) and (ii) together; (iv) $3\times 3$ convolution with stride of $(2,2)$ and 128 output maps; (v) $3\times 3$ convolution with stride of $(1,1)$, padding of $(1,1)$ and 128 output maps; (vi) skip connection by a $3\times 3$ convolution with stride of $(2,2)$ and 128 output maps with (iii) as input, and adding the results with (v);
(vii) $3\times 3$ convolution with stride of $(2,2)$ and 256 output maps; (viii) $3\times 3$ convolution with stride of $(1,1)$, padding of $(1,1)$ and 256 output maps; (ix) skip connection by a $3\times 3$ convolution with stride of $(2,2)$ and 256 output maps with (vi) as input, and adding the results with (viii);
(x) $3\times 3$ convolution with stride of $(2,2)$ and 512 output maps; (xi) $3\times 3$ convolution with stride of $(1,1)$, padding of $(1,1)$ and 512 output maps; (xii) skip connection by a $3\times 3$ convolution with stride of $(2,2)$ and 512 output maps with (viii) as input, and adding the results with (xi).
We build unidirectional connections after each convolutional layers in the plaintext branch.
After two branches, we have a 2-layer feature integration: In the first layer, we have two fully connected layers with 256 plaintext outputs and 256 ciphertext outputs, respectively; in the second layer, we have fully connected layer with 10 ciphertext outputs.

\begin{figure}[!tb]
\vskip 0.1in
\centering
\includegraphics[width=0.8\linewidth]{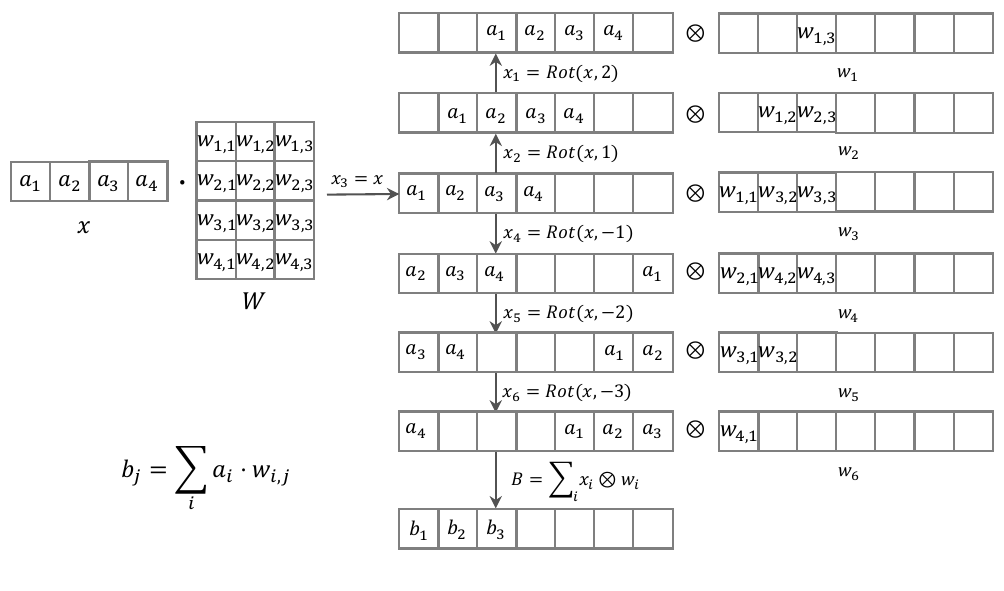}
\caption{Homomorphic vector-matrix multiplication.}
\label{fig:linear}
\vskip -0.1in
\end{figure}

\section{Implementation Details}
\label{app:imp}
In this section, we will introduce our implementation for bi-CryptoNets with CKKS scheme in more details.
A privacy-preserving neural network mainly consists of three types of layers: fully connected layer, convolutional layer and pooling layer.

\subsection*{Homomorphic fully connected layer}
For an input $\x\in\R^{n_1}$, weight matrix $W\in\R^{n_1\times n_2}$ and bias vector $\bm{k}\in\R^{n_2}$, the fully connected layer can be written as
\[
    y=\sigma(\x^TW+\bm{k})\ ,
\]
where $\sigma(\cdot)$ is the activation function.

As shown in Figure~\ref{fig:linear}, we can implement vector-matrix multiplication by homomorphic rotations and multiplications. The multiplication between $\x$ and $W$ can be written as
\[
    b_j=\sum_{i=1}^{n_1}a_i w_{i,j}\ \text{ for } \ j\in[n_2]\ .
\]
First, we transform the plaintext weight matrix $W$ into a group of vectors $w_i$, for $i\in[n_1+n_2-1]$. Then we perform rotations on $\x$ from $n_2-1$ to $1-n_1$ as
\[
    \x_i=Rot(x,n_2-i)\ \text{ for } \ i\in[n_1+n_2-1]\ .
\]
Next, we homomorphically multiply $x_i$ with $w_i$ respectively, and sum the results together. To save calculation depth, we can perform summation in pairs. This would only lead to $\lceil\log(n_2)\rceil$ of operation depth.

Finally, after adding bias $\bm{k}$ with prior result and perform activation function, we can get the output ciphertext of fully connected layers.

\begin{figure}[!tb]
\vskip 0.1in
\centering
\includegraphics[width=0.8\linewidth]{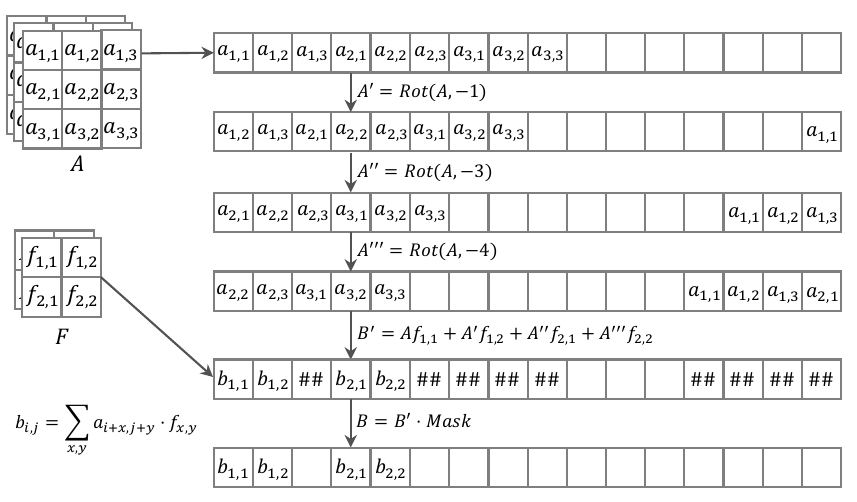}
\caption{Homomorphic convolution of one channel by HW packing with a 2x2 filter.}
\label{fig:conv}
% \vskip -0.1in
\end{figure}

\subsection*{Homomorphic convolutional layer}
Similar to homomorphic fully connected layer, we can implement the homomorphic convolutional layer by rotations and multiplications. As shown in Figure~\ref{fig:conv}, given an input channel $A\in\R^{h\times w}$ and convolution kernel $f\in \R^{h_c\times w_c}$, the convolution can be written as
\[
    b_{x,y}=\sum_{i,j}a_{x+i,y+j}f_{i,j}\ \text{ for }\ i\in[h],\ j\in[w]\ .
\]
For each input channel, it will be flattened into a one-dimensional vector and packed as a ciphertext. To perform the convolution on ciphertext, we recursively rotate it and multiply it by corresponding plaintext weight in the convolution kernel respectively. After adding them together, it will finally be multiplied by a mask vector, to only keep the values in the slots we need. We also can compute multiple convolution kernels for multiple input channels in parallel.

\subsection*{Homomorphic pooling layer}
Similar to prior works in private inference, we adopt sum pooling in our network. The implementation of pooling layer is similar to that of homomorphic convolutional layer. Nevertheless, we do not need to multiply weights in convolution kernels.

\subsection*{BHW Packing}

For inference, we could implement our bi-CryptoNets by packing method for speed on multiple images simultaneously, motivated by \cite{DathathriS0LLMM19}. We introduce the new BHW packing for our bi-CryptoNets without extra costs, and the basic idea is to pack multiple segments into one ciphertext for decomposed images. Specifically, we pack $n$ of $h\times w$ matrices into one ciphertext for a batch of $n$ images and their sensitive segments with channel $c$, height $h$ and width $w$. It requires $c$ ciphertext for $n$ images with our BHW packing.

\begin{wrapfigure}{r}{6.2cm}
\vskip -0.3in
\centering
\includegraphics[width=6cm]{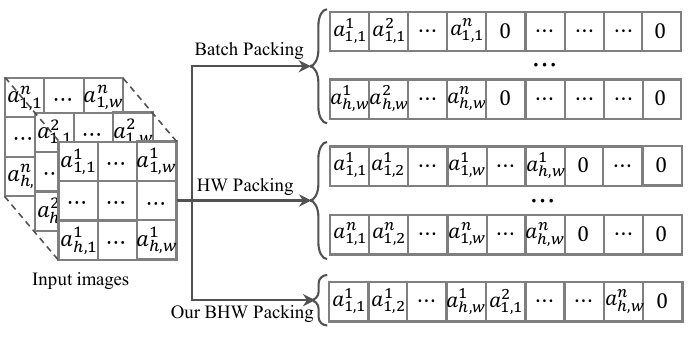}
\caption{A simple demonstration for three packing methods.}
\label{fig:pack}
\vskip -0.2in
\end{wrapfigure}

Batch packing \cite{Gilad-BachrachD16,abs-1811-09953} presents an efficient method for a large number of images, to pack the same pixels in a batch of images into one ciphertext. Another relevant HW packing tries to pack one channel of pixels in an image into one ciphertext \cite{DathathriS0LLMM19,LeeLLK0NC22}, which could greatly reduce HE computations and inference latency for single image.

Figure~\ref{fig:pack} presents a simple demonstration for three packing methods. Our BHW packing could achieve low inference latency as that of HW packing over one single image, and take comparable inference latency as that of batch packing on multiple images of smaller batch size. This is because our BHW packing method could pack more segments into one ciphertext for the decomposed images, and only small sensitive segments are required to be encrypted.

\begin{table*}[t]
\caption{The experimental comparisons on \textsf{CIFAR-100}, and abbreviations are similar to  Table~\ref{table2}.}\label{app:table3}\vspace{-2mm}
\begin{center}
\begin{small}
\begin{tabular}{lcccccccccc}
\toprule
\scriptsize Scheme & \scriptsize  HEOPs & \scriptsize  Add$_\text{CC}$ & \scriptsize  Mul$_\text{PC}$ & \scriptsize  Act$_\text{C}$ & \scriptsize  Latency(s) & \scriptsize  Amortized Latency(s) & \scriptsize  Acc(\%)\\
\midrule
\scriptsize{VDSCNN \cite{LeeLLK0NC22}} & \scriptsize  18K & \scriptsize  8K & \scriptsize  9K & \scriptsize  752 & \scriptsize  3942 & \scriptsize  3942 & \scriptsize  69.43\\
\scriptsize{Backbone CNN-11} & \scriptsize  1.0M & \scriptsize  493K & \scriptsize  521K & \scriptsize  246 & \scriptsize  1829 & \scriptsize  229 & \scriptsize  66.30 \\
\midrule
\scriptsize{\textbf{bi-CryptoNets with CNN-11}} & \scriptsize  709K & \scriptsize  341K & \scriptsize  368K & \scriptsize  246 & \scriptsize  1588 & \scriptsize  49 & \scriptsize  68.78\\
\scriptsize{\textbf{bi-CryptoNets with VGG-16}} & \scriptsize  3.4M & \scriptsize  1.5M & \scriptsize  1.9M & \scriptsize  1.1K & \scriptsize  2969 & \scriptsize  93 & \scriptsize  72.35\\
\scriptsize{\textbf{bi-CryptoNets with ResNet-18}} & \scriptsize  15.5M & \scriptsize  6.9M & \scriptsize  8.6M & \scriptsize  2.8K & \scriptsize  6765 & \scriptsize  212 & \scriptsize  73.33\\
\bottomrule
\end{tabular}
\end{small}
\end{center}
\end{table*}
\section{Results on \textsf{CIFAR-100}}\label{app:100}
In this section, we will present our experiment results on \textsf{CIFAR-100} dataset. As shown in Table~\ref{app:table3}, our bi-CryptoNets can achieve similar results on \textsf{CIFAR-100} as on \textsf{CIFAR-10} dataset, since the network structure are similar to each other.  Our bi-CryptoNets with CNN as backbone can could also decrease the amortized latency by $4.67\times$ and improve accuracy by $2.48\%$ in comparison with backbone network. Our bi-CryptoNets with VGG-16 and ResNet-18 as backbones still can achieve better inference accuracy, comparable inference latency and smaller amortized latency.

\begin{table*}[t]
\caption{The accuracies(\%) on \textsf{MNIST}, \textsf{CIFAR-10} and \textsf{CIFAR-100} datasets.}\label{app:table}
% \vskip 0.1in
\begin{center}
\begin{small}
\begin{tabular}{|l|c|ccc|cc|}
\toprule
\multirow{2}{*}{ \scriptsize Models}  & \scriptsize  \textsf{MNIST}  &  \multicolumn{3}{c|}{\scriptsize\textsf{CIFAR-10}} &  \multicolumn{2}{c|}{\scriptsize\textsf{CIFAR-100}} \\
 & \scriptsize  CNN-3 & \scriptsize  CNN-11 & \scriptsize  VGG-16 & \scriptsize  ResNet-18 & \scriptsize  VGG-16 & \scriptsize  ResNet-18\\
\midrule
\scriptsize{Backbone network without decomposition} & \scriptsize  $99.21$ & \scriptsize  $90.99$ & \scriptsize  $93.42$& \scriptsize  $94.30$& \scriptsize $72.35$ & \scriptsize  $74.00$ \\
\midrule
\scriptsize{bi-CryptoNets (Convolution resizing)}  & \scriptsize  $98.96$ & \scriptsize  $89.69$ & \scriptsize  $92.24$ & \scriptsize  $92.05$ & \scriptsize  $71.51$ & \scriptsize  $70.79$ \\
\scriptsize\textbf{bi-CryptoNets (Cropping resizing)} & \scriptsize  $\mathbf{99.15}$ & \scriptsize  $\mathbf{90.30}$ & \scriptsize  $\mathbf{93.27}$& \scriptsize  $\mathbf{93.91}$ & \scriptsize  $\mathbf{72.23}$ & \scriptsize  $\mathbf{73.33}$\\
\bottomrule
\end{tabular}
\end{small}
\end{center}
% \vskip -0.1in
\end{table*}

\section{Unidirectional Connections}
In Section~\ref{sec:uni}, we use cropping as resizing function for simplicity. Another way is to use a convolutional layer with strides as resizing function. In this section, we will show that cropping is a more effective way.

Similar to the skip connections of ResNet, we can simply use a convolutional layer as resizing functions. For example, if both the height and width of the sensitive segment is a half of that of insensitive segment, we can use a $3\times 3$ convolution with stride of $(2,2)$ as resizing functions. However, such approach has following disadvantages.

First, extra convolutional layers would introduce more parameters, which would make the network harder to train, and it would also lead to larger computation overhead. Second, when using cropping as resizing functions, it can bring representation from the plaintext branch at similar depth to the corresponding layers in ciphertext branch, whereas convolutions will bring higher orders representation to the layers in ciphertext branch. Intuitively, it is more natural to adding representations from similar depth together, compared with adding a high-order representation with low-order representation.
Finally, cropping is an easier and more flexible way for resizing feature maps with various shapes, compared with convolution. Therefore, we simply use cropping as resizing functions in this paper.

Moreover, to compare the effectiveness of two resizing functions, we conduct experiments on \textsf{MNIST}, \textsf{CIFAR-10} and \textsf{CIFAR-100} with different network structures. We replace the cropping resizing function in our bi-CryptoNets with $3\times 3$ convolution by stride of $(2,2)$. We train both networks with feature-based knowledge distillation, and compare their accuracy. As shown in table~\ref{app:table}, using cropping as resizing function can reach better accuracies in all datasets and with all variants of bi-CryptoNets structures, which is consistent with our analysis.

\section{Visualization for the effectiveness of feature-based knowledge distillation}\label{app:visual}
\begin{figure}[!tb]
\vskip 0.1in
\centering
\includegraphics[width=0.9\linewidth]{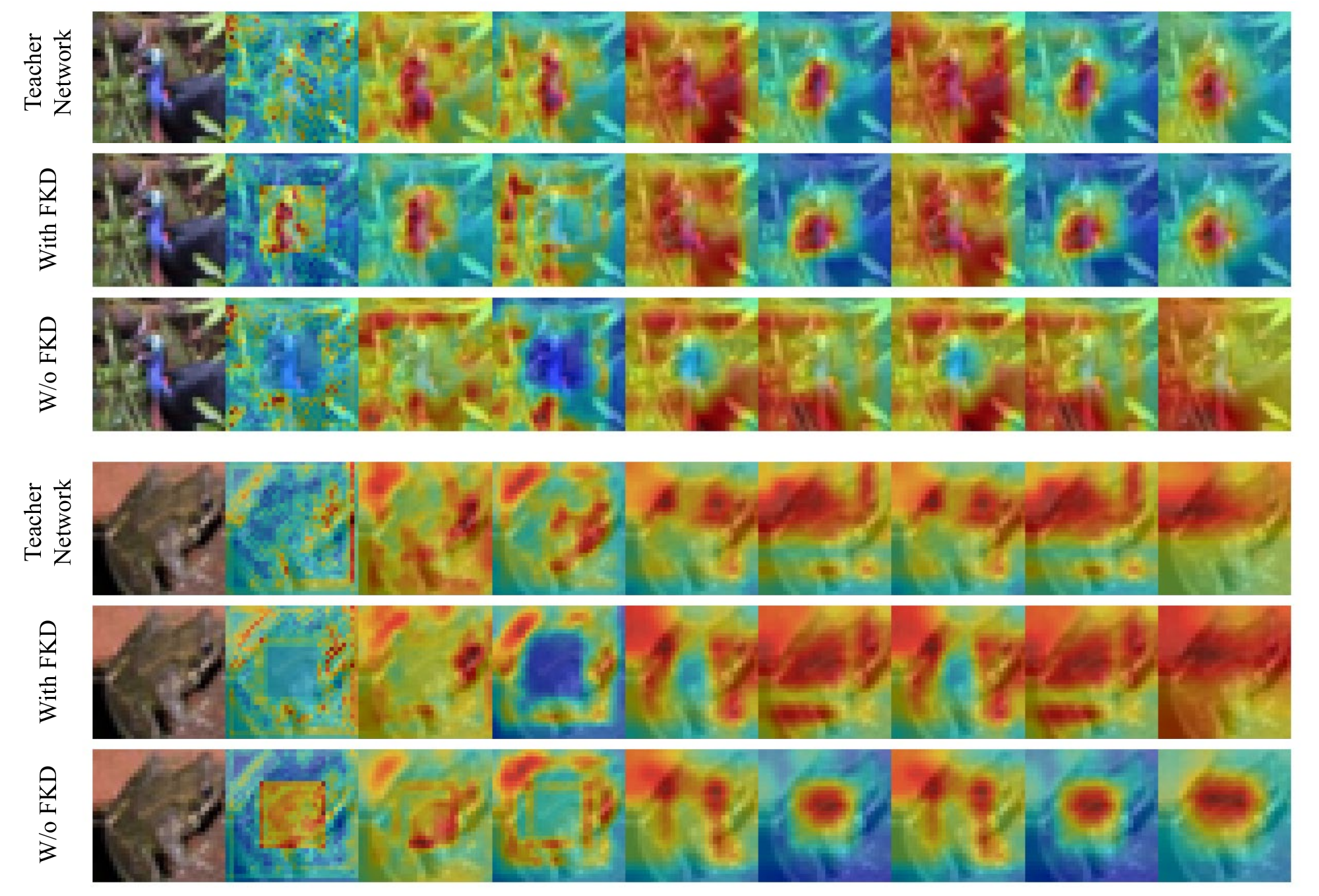}
\caption{Attention maps in different layers of teacher network (CNN-11), bi-CryptoNets with feature-based knowledge distillation and bi-CryptoNets without knowledge distillation on \textsf{CIFAR-10} images.}
\label{fig:v1}
\vskip -0.1in
\end{figure}

In this section, we will show the effectiveness of feature-based knowledge distillation for bi-CryptoNets by drawing the attention maps at each layer. We compare our bi-CryptoNets trained by feature-based knowledge distillation with the teacher network and the bi-CryptoNets trained without knowledge distillation. Figure~\ref{fig:v1} presents the attention maps of each layer for bi-CryptoNets that takes 11-layer CNN as backbone on \textsf{CIFAR-10} images, and Figure~\ref{fig:v1} presents the attention maps of each layer for bi-CryptoNets that takes VGG-16 as backbone. We can observe that the attention maps of bi-CryptoNets with feature-based knowledge distillation is very close to those of teachers, whereas bi-CryptoNets without knowledge distillation fails to capture the important patterns in the images, resulting the wrong predictions. Therefore, our feature-based knowledge distillation is an effective way to improve the performance of bi-CryptoNets, so that our bi-CryptoNets can achieve comparable performance as teacher network.

\begin{figure}[!tb]
\vskip 0.1in
\centering
\includegraphics[width=0.9\linewidth]{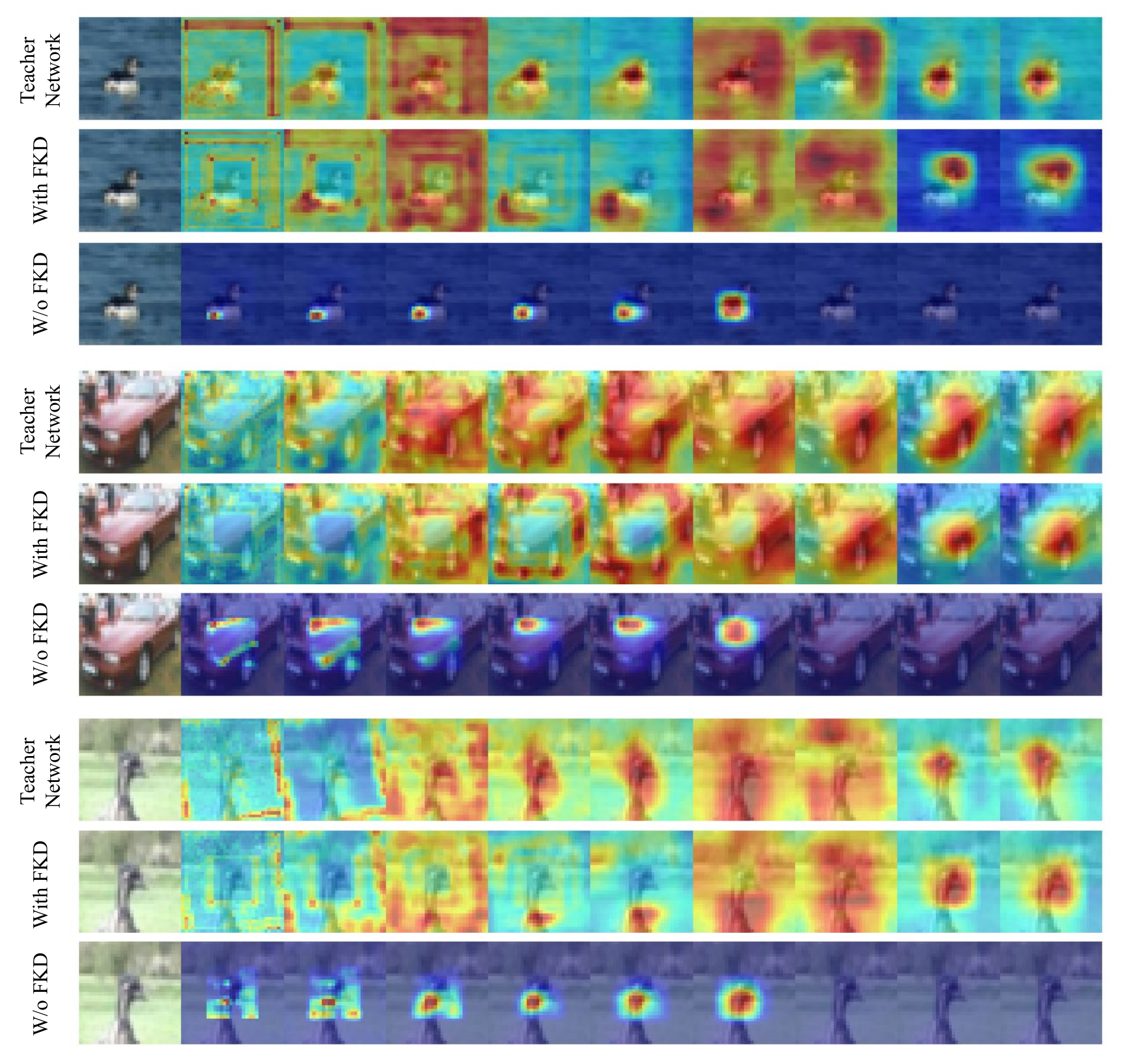}
\caption{Attention maps in different layers of teacher network (VGG-16), bi-CryptoNets with feature-based knowledge distillation and bi-CryptoNets without knowledge distillation on \textsf{CIFAR-10} images.}
\label{fig:v2}
\vskip -0.1in
\end{figure}

\end{document}